\documentclass[%
 aip,
 amsmath,amssymb,
 reprint,%
]{revtex4}

\usepackage{graphicx}
\usepackage{dcolumn}
\usepackage{bm}

\usepackage{multirow}

\usepackage[utf8]{inputenc}
\usepackage[T1]{fontenc}
\usepackage{mathptmx}
\usepackage{etoolbox}

\makeatletter
\def\@email#1#2{%
 \endgroup
 \patchcmd{\titleblock@produce}
  {\frontmatter@RRAPformat}
  {\frontmatter@RRAPformat{\produce@RRAP{*#1\href{mailto:#2}{#2}}}\frontmatter@RRAPformat}
  {}{}
}%
\makeatother
\begin{document}

\preprint{AIP/123-QED}

\title[Analysis of chaotic dynamical systems with autoencoders]{Analysis of chaotic dynamical systems with autoencoders}
\author{N.  Almazova$^1$}
\author{G. D. Barmparis$^{1, 2,*}$}
\email{barmparis@physics.uoc.gr}
\author{G. P. Tsironis$^{1, 2}$}
\affiliation{$^1$National University of Science and Technology MISiS,  Leninsky prosp. 4, Moscow 119049, Russia}
\affiliation{$^2$Institute of Theoretical and Computational Physics and Department of Physics, University of Crete, P.O. Box 2208, 71003 Heraklion, Greece}

\date{\today}

\begin{abstract}
We focus on chaotic dynamical systems and analyze their time series with the use of autoencoders, i.e., configurations of neural networks that map identical output to input. This analysis results in the determination of the latent space dimension of each system and thus determines the minimal number of nodes necessary to capture the essential information contained in the chaotic time series. The constructed chaotic autoencoders generate similar maximal Lyapunov exponents as the original chaotic systems and thus encompass their essential dynamical information.
\end{abstract}

\maketitle

\begin{quotation}
The quantitative study of complex systems may be assisted by the use of methods derived from neural networks. The analysis of chaotic time series presents an applied challenge towards determining the features involved and particularly the effective complexity they include. This work shows that the use of chaotic autoencoders made of neural networks leads to essential information on the time series and the chaotic systems they arise from. The autoencoders we construct map identically the input chaotic time series to the output. The process passes through the latent space that acts as an information bottleneck for this reconstruction. We estimate the dimension of the latent space numerically and find it smaller than possible embedding dimensions of the reconstruction. The dimension of the latent space hints towards the complexity of the dynamical system underlying the time series. Furthermore, we show that the constructed chaotic autoencoders produce maximal Lyapunov exponents that are very close to those obtained directly from the equations of motion of the chaotic systems. These features show that the constructed chaotic autoencoders are faithful representations of the time series generating dynamical systems, and through their latent space dimension, we obtain an estimate of their complexity.
\end{quotation}

\section{\label{sec:intro}Introduction}
Nonlinear dynamical systems, either continuous or discrete, may produce chaos, i.e.  irregular evolution similar to probabilistic, stochastic  motion.  Continuous autonomous chaotic systems are described by at least three coupled ordinary nonlinear differential equations that, when solved numerically, result in trajectories with irregular features.  In general, non-Hamiltonian cases,  the trajectory may fall on a strange attractor that is a chaotic yet distinct structure in the system phase space.  Understanding the complex dynamics in phase space is significant especially in the mathematical analysis of the inverse problem.  In the latter we are given a time series of discrete data and we want to find whether this set derives from a chaotic dynamical system or it is purely stochastic \cite{Anishchenko, Abarbanel, Kantz,Bradley}.  This is a significant problem that stems from the need to understand experimental data as well as other data that are readily available in our times.  In this problem, we are given a time series of data and we are interested in reconstructing the phase space of the associated dynamical system \cite{Abarbanel2, Deyle, Verdes}.  The methods that have been developed over the last thirty or more years work quite well and the area is mature.  Nevertheless, since these methods are necessarily cumbersome and involve additionally empirical inference it would be interesting to experiment with the use of methods derived from Artificial Intelligence (AI).  More specifically, Machine Learning (ML) provides with an ensemble of well tested methods that may assist or even surpass traditional methods of analysis of chaotic systems \cite{Goodfellow, Jordan}.  The basic aim of the present work is to use unsupervised learning methods and attempt to reconstruct the dynamical features of the data without relying on the traditional methods.  

We propose an approach that employees autoencoders,  a ML tool that is used widely,  as a dimensionality reduction technique.  Autoencoders are implemented with neural networks and are in effect unit projection operators where the output is as close as possible to the input. In the process of this unit projection, data pass through the latent space where they suffer dimensionality reduction.  The effective dimension of the latent space gives the degree of possible data compression and thus eliminates original dimension redundancy.  Since the neural network uses fewer nodes in latent space than the number of input nodes we accomplish  compression of the original data. This packing allows detecting the basic information contained in the data. This compression technique involves data regularization in latent space that is responsible for the selection of the optimal number of operating nodes.  This strategy allows us to find the essential number of nodes for the trained network that recapture the complexity of the original time series.

Once the minimal configuration in latent space is found, we need to compare the chaotic autoencoder with the original system.  This is accomplished through the comparison of the corresponding maximal Lyapunov exponents, i.e. that of the original trajectories of the system with the one calculated from the reconstructed trajectories generated through the autoencoder.  The Lyapunov exponents are a general attribute of nonlinear dynamical systems as they allow to distinguish the local stability and invariant sets in a particular direction of attractors. The largest Lyapunov exponent detects the mean divergence of nearest neighbors in two initial nearby trajectories. The positive value of the exponent defines the degree of chaos in a particular system. There exist different methods for finding the largest Lyapunov exponent \cite{Wolf,  Sano, Rosenstein, Eckmann, Parlitz}. In this work we use the largest Lyapunov exponent as a measure of the level of chaos resulting  from neural network chaotic autoencoders. 

Machine Learning techniques are used in areas far beyond Computer Science. Neural networks (NN) are applied for a wide range of tasks that permit solving problems in engineering as well as science while deep learning expands the ability of generalization, interpolation, etc \cite{Jordan, Nielsen}.  Artificial intelligence helps to solve  tasks through appropriate data handling but nevertheless they cannot give deep insight to the solution processes they use.  They operate by allowing the computers to learn the features of data and subsequently understand the hierarchies involved.  Several Machine Learning techniques, like the reservoir computing \cite{Ott,  Budhiraja},  the long short-term memory (LSTM) \cite{Vlachas} and the O-LSTM \cite{Neofotistos}, have been used to investigate the time evolution of complex systems when the time series is known \cite{Brunton,  Dercole,  Lusch}. Autoencoders are also used in the physical systems due to their main advantage of dimensionality reduction \cite{Goodfellow,  Lopez}.  An autoencoder is a neural network that maps the input data to the output \cite{Kramer}.  It presents a method of unsupervised learning with the compression of data occurring in the latent space that forms a kind of information bottleneck, due to the smaller number of nodes it includes.  The target data for an autoencoder are the input data. Traditionally,  autoencoders are used as dimensionality reduction techniques and salient features of learning. The construction of an autoencoder includes two main parts. The first part is an encoder function, $h(x)$, that compresses the input information, $x$, while the second part is the decoder, $g(h)$ takes as input the compressed information and by decompressing it reconstructs the initial input of the encoder.  The output function can be written as $\hat{x} = g \big( h(x) \big)$,  where the input sequence should be learned ``perfectly". In other words, the autoencoder attempts to copy the data to the output while first they pass through a compression bottleneck.  The learning results are obtained through a minimization of the appropriate loss function. 

In this article we focus on the construction of chaotic autoencoders.  In the next section, we detail features of dynamical systems such as phase space reconstruction and largest Lyapunov exponents and report on the methods for their calculation. Subsequently, we define the chaotic autoencoders for three prototypical dynamical systems and show how the information compression operates. We then show how the Lyapunov exponents are evaluated in the chaotic autoencoders, and finally, we conclude.

\section{\label{sec:systems}Description of the chaotic systems}

A deterministic chaotic signal behaves as something intermediate between a regular and a stochastic motion.  The nonlinear aspect of this behavior of the signal makes difficult the prediction and classification of the physical system, from which the motion comes from \cite{Deyle, Sauer, Hilborn, Lorenz}. A chaotic motion produces specific structures in phase space. Systems of differential equations with three or more degrees of freedom may show chaos in their evolution in real or phase space. The number of degrees of freedom of a system describes the number of the necessary independent variables to specify the dynamical state of the system. Chaotic systems are described through their time-evolution equations, the values of their parameters, and the selected initial conditions. We may express the ordinary differential equations guiding the evolution of a state using a variable $x(t) = [x_1(t), x_2(t), \ldots,  x_i(t)]$ where $x_i (t)$ is the variable describing the $i-th$ degree of freedom at time, $t$,  as:
\begin{equation}
\frac{dx(t)}{dt} = f(x(t))
\end{equation}
where, $f$, describes the nonlinear function that characterizes the equations of motion of the system.  Given the system of differential equations guiding a dynamical system, we can solve them numerically and follow its evolution in time. Many dynamical systems have been investigated using this methodology. Here, we focus on prototypical systems well known in the literature of chaotic dynamics, viz. the R\"ossler system and the Lorenz system. Since we want to have a more general idea of the effectiveness of our method, we use both the Lorenz63 system that contains three degrees of freedom and the Lorenz96 one with multiple dependent variables and external forcing.  We will describe these systems below.

\subsection{Delay-coordinate embedding}

A physical system can only be measured discretely in time, with a given time step,  and thus it can be represented as a time series of data. In the case of a chaotic system,  we can use several methods to determine the features of the chaotic data. One of these methods is the time-delay embedding or reconstruction scheme \cite{Abarbanel2,  Hilborn,  Broomhead}. This method allows for the recovery of the entire multidimensional observation in state-space based on the time series of just one of its variables. Let us assume that a time series, $X$, represents a trajectory of a physical system recorded with a fixed time step as an $N$-points time series. The reconstructed trajectory, $X$, can be represented as an $(M \times m)$ matrix,
\begin{equation}
X = \big[X_1, X_2, \ldots,X_M\big]^T,
\end{equation}
where,  $M = N - (m-1) \tau$,  $m$ is an embedding dimension,  $\tau$ is a time lag (delay) and $X_i$ is a vector representing the state of the system at discrete time $i$, given by:
\begin{equation}
X_i = \big[ x_i,  x_{i+\tau}, \ldots, x_{i + (m-1)\tau} \big],
\end{equation}
with, $ x_i$, being the, $i-th$, data point of the time series.  In order to find the embedding dimension for a dynamical system,  one can implement the Takens theorem \cite{Takens}. Takens introduced the idea that a complete part of dynamics with basic dimension $d_s$ might be observed in an embedding space if the latter has $2d_s + 1$ or higher dimensions. There are a number of practical implementations of Takens theorem although these methods are not necessarily universal \cite{Kantz, Packard, Rosenstein2}.  In the present work, we use a time lag equal to one and an appropriate embedding dimension that satisfies the Takens theorem.

\section{Methodology}

\subsection{Data Preparation}

In this work, we analyze three chaotic systems, viz.  that of R\"ossler,  Lorenz63, and Lorenz96. For the sake of brevity, we present the details of each system in Table \ref{table:params}.  We integrate the differential equations of each system by means of the 4$^{th}$ order Runge-Kutta method, using a fixed time step equal to 0.005, for 300000 time steps. The first 50000 time steps of each solution were removed as transient points.  We use the x-coordinate of the R\"ossler and the Lorenz63 systems, and the 19$^{th}$ coordinate (out of forty) of the Lorenz96 as the training sequences. The 250000 time steps of each training sequence are normalized between zero and one and split into training (80\%) and testing (20\%) sets.  The training and testing sets are split into smaller pieces using a sliding window of a given size,  $W$. 

\begin{table}[h]
\centering
\begin{tabular}{c|c|c|c}
System & Equations & Parameters \\
\hline
R\"ossler & & \\
\multirow{4}{*}{ \includegraphics[height=1.5cm]{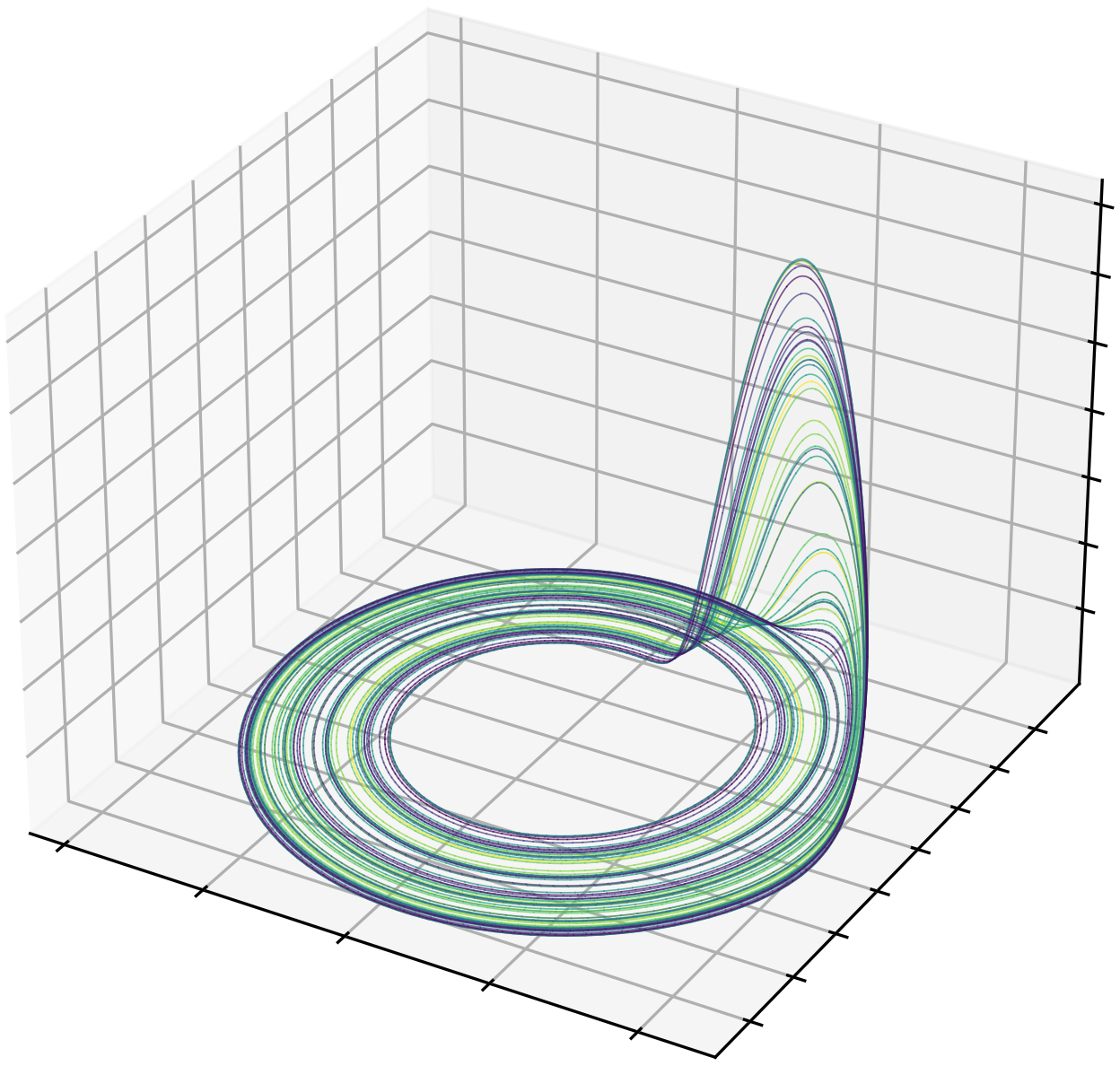} } & $\frac{dx}{dt} = y - z $ & a = 0.1 \\
                                              & $\frac{dy}{dt} = x + ay $ & b = 0.1 \\
                                              & $\frac{dz}{dt} = b + z(x - c)$ & c = 14 \\
                                              & & \\
\hline
Lorentz63 & & \\
\multirow{4}{*}{ \includegraphics[height=1.5cm]{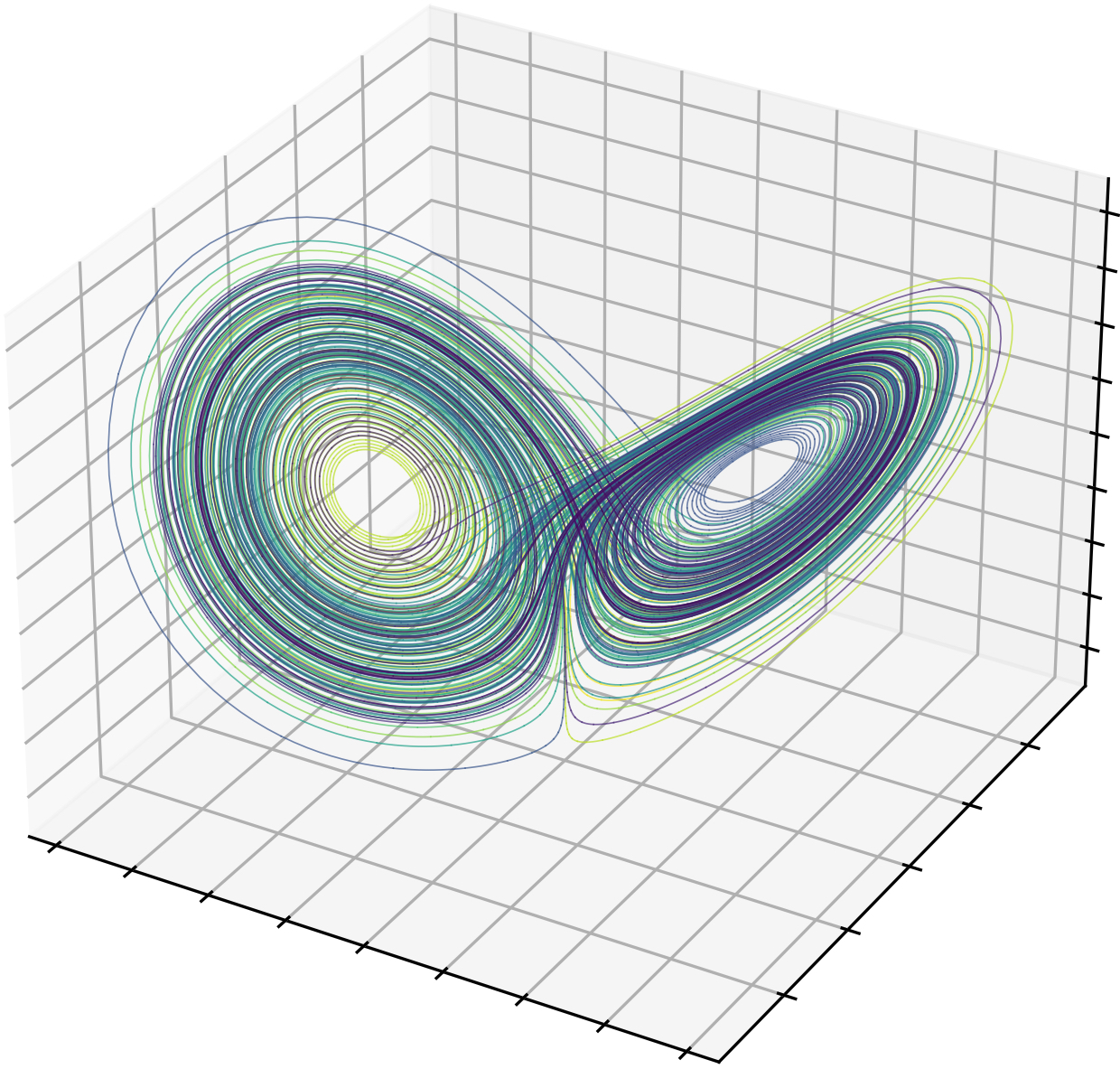} } & $\frac{dx}{dt} = \sigma (y - x)$ & $\sigma$ = 10 \\
                                                & $\frac{dy}{dt} = x(\rho - z) - y$ & $\rho$ = 28 \\
                                                & $\frac{dz}{dt} = xy - \beta z$ & $\beta$ = 8/3 \\
                                                & & \\
\hline
Lorentz96 & & \\
\multirow{4}{*}{ \includegraphics[height=1.5cm]{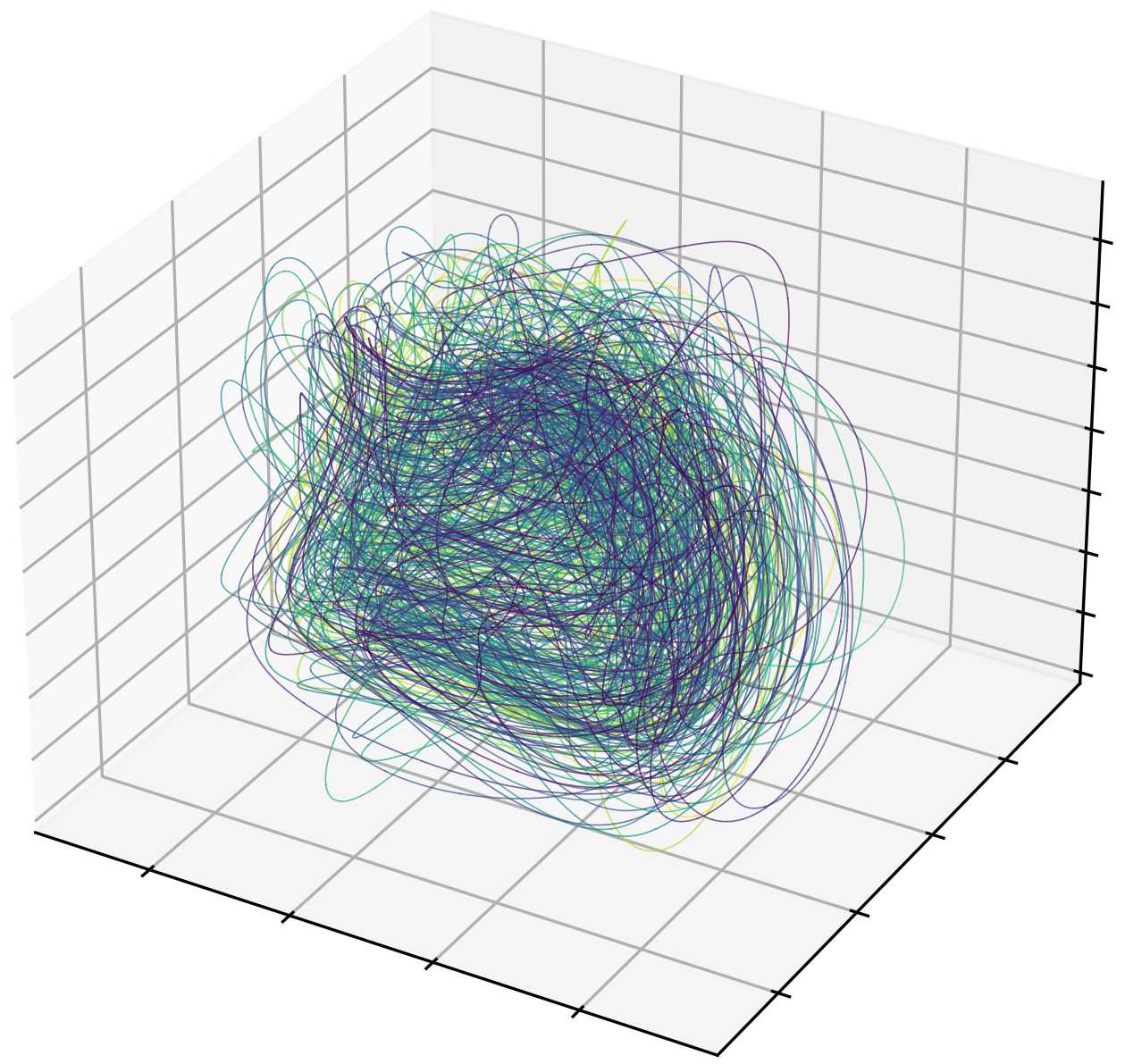} } &  & N = 40 \\
                                                & $\frac{dx_i}{dt} = (x_{i+1} - x_{i-2}) x_{i-1}  - x_i + F$ &  \\
                                                &  & $F$ = 8.15 \\
                                                &  & \\
\hline
\end{tabular}
\caption{A graphical representation, the equations and the parameters of each one of the chaotic systems in consideration.}
\label{table:params}
\end{table}

\newpage

\subsection{Machine Learning: Setting up the chaotic autoencoder}

\begin{figure}[h]
\includegraphics[width = 14cm]{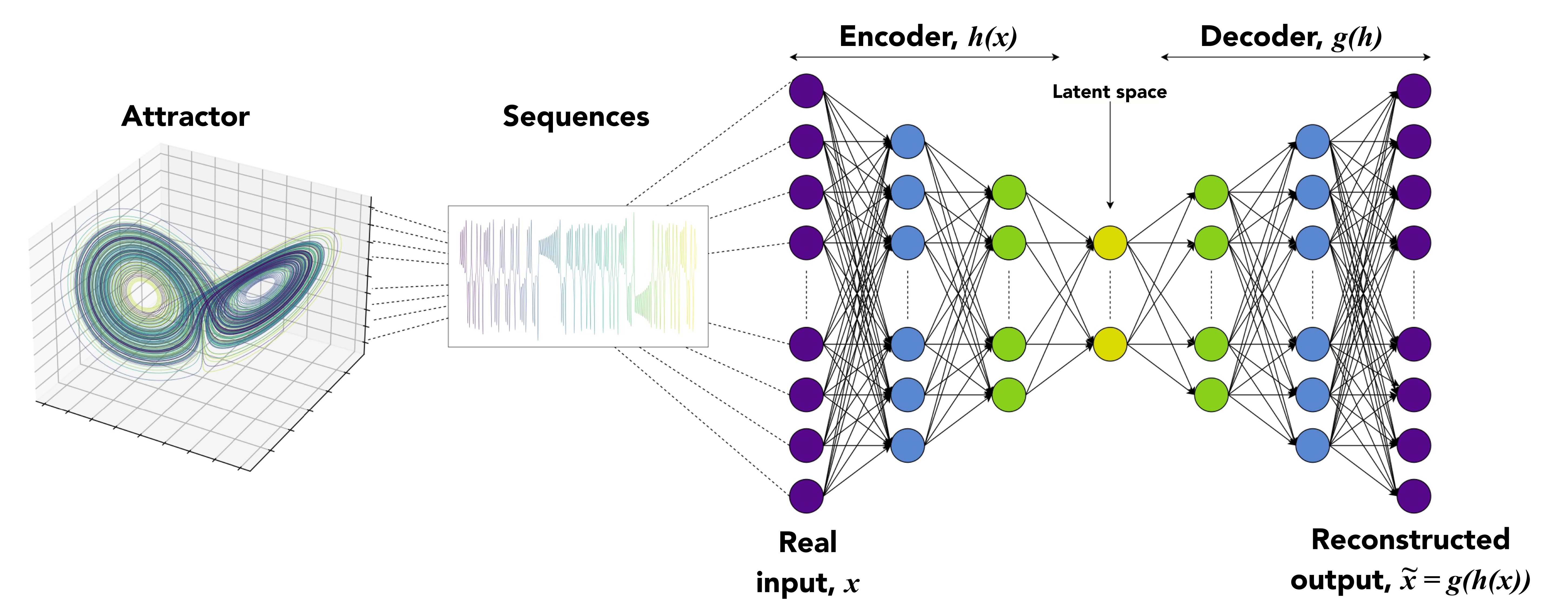}
\caption{A schematic representation of the proposed model.  The trajectories of a given attractor are transformed into input sequences, $x$, that pass through an encoder model, $h(x)$,  get compressed into the latent space and pass through the decoder, $g(h)$, to reconstruct the original information, $\hat{x} = g(h(x))$.}
\label{autoencoder}
\end{figure}

A stacked autoencoder composed of two symmetric networks, the encoder, $h(x)$,  and the decoder, $g(h)$, is used to reconstruct each one of the chaotic systems in this work.  The encoder consists of an input layer with 30 nodes, i.e., equal to the window size, $W$, used to split the input sequences. Then, two fully connected hidden layers of 22 and 15 nodes with a Sigmoid and ReLU activation function, respectively, guide the model to the latent space. The latent space is represented by a fully connected layer of 10 nodes and a ReLU activation function.  L1 regularization (Lasso Regression) is applied to the output of the nodes of the latent space to force nodes with less information to zero.  The L1 regularization applied to the output of a layer adds to the loss function a value proportional to the regularization parameter, $\alpha$, to shrink nodes with less information. The regularization parameter, $\alpha$, is used as a hyper-parameter of the model.The larger the regularization parameter, the more we constrain the autoencoder and thus we increase the error in the reconstructed sequence. The smaller the regularization parameter, the larger the active part of the latent space and thus we decrease the error in the reconstructed sequence. After finishing with the training process, we statistically analyze the contribution of the number of active nodes in the latent space to accuracy of the the reconstructed sequence. Nodes that output values smaller than the regularization parameter contain non-important information and their contribution to the reconstructed sequence is negligible. Nodes with values larger than the value of $\alpha$ are counted as non-zero nodes with significant contribution to the output.  The decoder network is constructed as the mirror image of the encoder one.  The autoencoder model is trained for 7500 epochs with a batch size equal to 32, using the Adam stochastic optimization method \citep{ADAM} with a learning rate of 0.001, while all its weights are initialized using the Xavier uniform initializer \cite{Xavier},  as implemented in Tensorflow/Keras \cite{TF, Keras}.  In Fig. (\ref{autoencoder}), we present a schematic representation of the proposed model. 

\begin{figure}[h]
\includegraphics[width = 8cm]{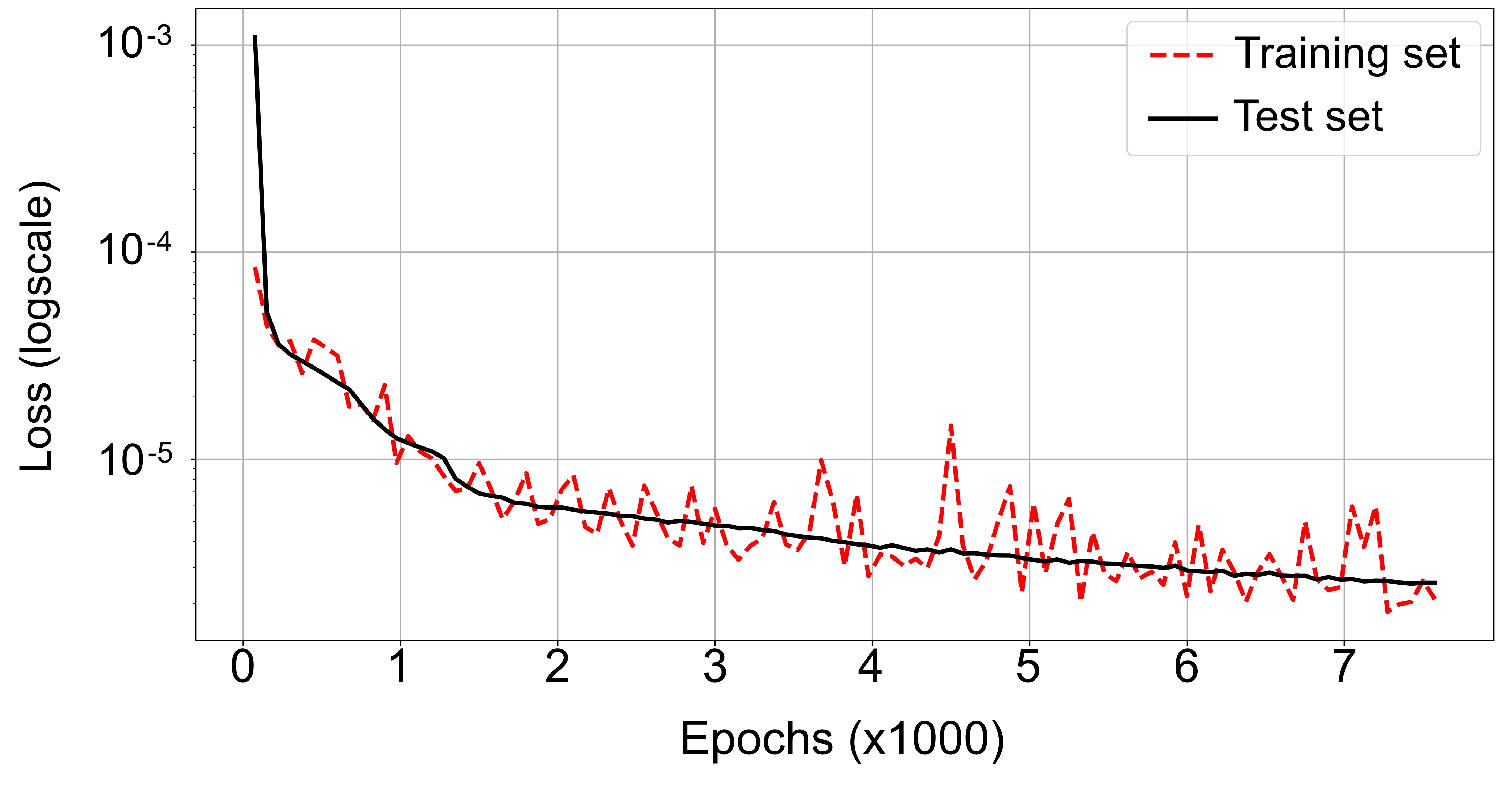}\\
\caption{The evolution of the training (red dashed line) and the test set (solid black line) loss in logarithmic scale, as a function the number of epochs of training for the Lorenz63 system. }
\label{losses}
\end{figure}

As mentioned in the Introduction section, an autoencoder is a neural network that acts as an identity operator and maps the input data ideally to identical output data.  In performing this operation through the latent space, it must learn through minimizing a loss function. The simplest example of a loss function is that of the mean squared error (MSE) , i.e.
\begin{equation}
MSE(x, \hat{x}) = \frac{1}{W}\sum_{k=1}^{W} \big( x_k - \hat{x}_k \big)^2,
\end{equation}
where, $W$ is the length of the input sequence, $x$, represents the real - input data (a set of $x_k$ real samples),  $\hat{x} = g(h(x))$ represents the reconstructed - output data (a set of $\hat{x}_k$ reconstructed samples) of a general ML system, while the loss function "punishes" the difference between $x$ and $\hat{x}$, i.e., the real and reconstructed data.  Here, to train the autoencoder we additionally apply the sparsity penalty (L1 regularization), $\Omega(h) = \alpha |h(x)|$, on the output of the encoder, $h(x)$, (after the activation function), where, $\alpha$, is the regularization parameter. Then the loss function, $L$, is represented as:
\begin{equation}
L \big( x,  \hat{x} \big) = MSE + \Omega(h(x)).
\end{equation}
This feature permits the autoencoder to limit the number of nodes at the latent space to a minimum required number, that can be used to reconstruct the original signal correctly. This type of regularization forces the participation of nodes in latent space to be zero when their contribution to the output of the latent space is less than the regularization parameter.  Given the fact that we want to investigate the performance of the model as a function of the regularization parameter, $\alpha$,  in order to demonstrate the effect of the regularization and find the minimum number of nodes that can correctly reconstruct the input sequences,  we fix all the other parameters of our model and we train it only once.  Thus, we can compare the results for the different values of the regularization parameter.  In Fig. \ref{losses}, we present the evolution of the loss function during the training process for the training and the test set of the Lorenz63 system.

\subsection{Determining $\alpha$-dependent latent space dimension}

The regularization parameter, $\alpha$, controls the size of the latent space and plays a crucial role in determining the number of the important nodes in the latent space. In our approach, we define as important the nodes in the latent space that output a value larger than the regularization parameter, $\alpha$. Thus, we calculate the average of non-zero nodes in the test set as a function of $\alpha$. The results for the three systems in consideration are shown in Fig. \ref{latentSpaceVsAlphaValue}.

\subsection{Largest Lyapunov Exponent as a Metric}

The Lyapunov exponents are an essential concept of nonlinear dynamics and present specific features of chaotic systems. Two trajectories of the attractor initiated at points displaced infinitesimally from each other will diverge exponentially if the attractor is chaotic. The exponential discrepancy is described by the largest Lyapunov exponent (LLE). This definition can be extended to the spectrum of Lyapunov exponents. The LLE of a system can be calculated using two approaches. The first approach uses a Jacobian matrix and can be applied when the mathematical model describing the evolution of the system is known. This method gives the Lyapunov spectrum. The second one can be used to estimate the LLE of a time series by applying the definition of the local divergence on a segment of a given trajectory.  Estimating the LLE from a time series faces several numerical difficulties, like dealing with the noise in the data \cite{Wolf, Rosenstein, Eckmann}. The basic definition of the LLE, $\lambda_1$, can be written as:  
\begin{equation}
d(t) \approx C e^{\lambda_1 t}
\end{equation}
where, $d(t)$,  is an average distance at time, $t$ and $C$ is a constant (usually normalized by the initial separation of neighbors).  Here we use the methodology proposed by Rosenstein,  Collins and De Luca \cite{Rosenstein}, where we divide the time, $t$, into steps of $\Delta t$, and assume that at time, $t = i \times \Delta t$, the distance of the $j^{th}$ pair of nearest neighbors,  $d_j(i)$, diverges approximately at a rate given by the LLE:
\begin{equation}
d_j(i) \approx C_j e^{\lambda_1 i \Delta t},
\label{LLE_approx}
\end{equation}
where, $C_j$ is the initial distance of the selected pair of points. By taking the logarithm of Eq. (\ref{LLE_approx}) we have that:
\begin{equation}
\ln d_j(i) \approx \ln C_j + \lambda_1 (i \Delta t).
\label{ln_LLE_approx}
\end{equation}
Equation (\ref{ln_LLE_approx}), represents a set of $M$, (all the values of index $j$) approximately parallel lines that can be fitted using least squares to calculate an averaged value and its standard deviation of the LLE. 

In this work,  we compare the average value and its standard deviation of the LLE calculated using the original and the reconstructed trajectories.  The reconstructed dynamics obtained by the autoencoder depend on the parameter in L1 regularization - $\alpha$.  We report these findings and the dependence of LLE on the embedding dimension window in the results section.

\section{Results}
The basic results of this work consist on the evaluation of the effective size of the latent space of the chaotic autoencoder constructed for each of the three systems we considered.  The size of the latent space depends on the regularization parameter $\alpha$ and the window size of input sequence of the time series, $W$.  Once the autoencoder is trained we also compare the largest Lyapunov value it produces.

\subsection{Latent space size as a function of the regularization parameter, $\alpha$}

\begin{figure}[h]
\includegraphics[width = 16cm]{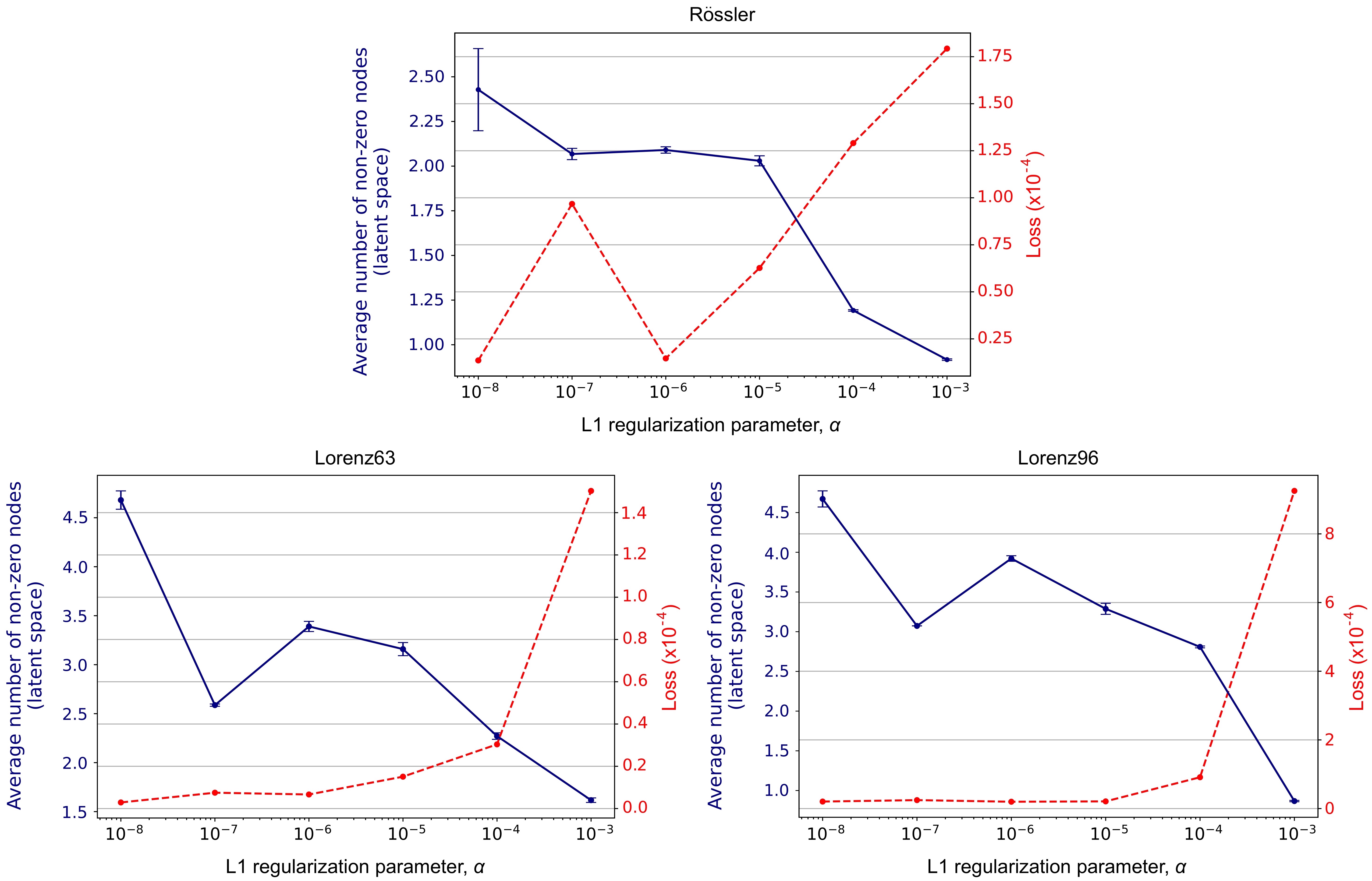}
\caption{The average number of non-zero nodes at the latent space (navy blue solid line), including the standard deviation of it (error bars) and the corresponding value of the loss function (red dashed line) of the test data as a function of the L1 regularization parameter, $\alpha$, for the three systems.}
\label{latentSpaceVsAlphaValue}
\end{figure}

We apply the methods outlined above on the three chaotic systems, presented in Table \ref{table:params}. In Fig.  \ref{latentSpaceVsAlphaValue}, we present the average number of non-zero nodes at the latent space (nodes with output value larger than $\alpha$) as well as the loss of the reconstructed sequence, with respect to the input sequence, as a function of the L1 regularization parameter, $\alpha$. 

We conclude that we find the same patterns for all studied systems, i.e., the loss function increases with the increment of the regularization parameter. Depending on which level of accuracy could be needed the number of nodes might change.  Especially,  we observe that for the R\"ossler system the autoencoder needs two nodes at the latent space to reconstruct the input sequence with error less than $10^{-4}$, while three to four nodes are needed in the case of the Lorenz systems. 

To further demonstrate the efficiency of the reconstructed trajectories by the autoencoder model we calculate the largest Lyapunov exponent for the input and the reconstructed sequences of each one of the regularization parameter values.  We calculate the LLE by creating two trajectories of each attractor. We choose a starting point for one of them and then we displace it by $10^{-7}$ for the second trajectory. We create the two trajectories by solving the equations of each attractor, and then we use the trained autoencoder of the corresponded system to reconstruct the trajectories.  In Table \ref{table:LLEsVsAlpha}, we present the LLE and its standard deviation in parenthesis for the input sequence, and the reconstructed sequence of each one of the values of the regularization parameter, $\alpha$. The results indicate that the reconstructed sequences created by models with $\alpha$ smaller or equal to $10^{-4}$ have the same characteristics as the sequences obtained by solving the deterministic system of equations.  The error in the average value of LLE is less than $\pm$ 2\%.

The obtained agreement between the system and chaotic autoencoder LLE is quite promising since it shows that the latter acts as an effectively linear system driven by a chaotic signal. It thus, demonstrates further that the chaotic autoencoder is a faithful reconstruction of the original system in the specific parameter regime.

\begin{table}[h!]
\centering
\begin{tabular}{c|c|c|c|c|c|c|c|c|c|c|}
& & \multicolumn{3}{c|}{R\"ossler} & \multicolumn{3}{c|}{Lorenz63} & \multicolumn{3}{c|}{Lorenz96} \\
\hline
Sequence & $\alpha$ & X & Y & Z & X & Y & Z & 19 & 20 & 21\\
\hline
Input & - & 1.73 (0.08) & 1.72 (0.05) & 1.74 (0.09) & 1.17 (0.30) & 1.17 (0.35) & 1.14 (0.18) & 3.02 (0.88) & 3.03 (0.35) & 3.06 (0.57) \\
\hline
\multirow{6}{*}{Reconstructed} & $10^{-3}$ & 1.76 (0.05) & 1.73 (0.02) & 1.77 (0.13) & 1.16 (0.42) & 1.13 (0.21) & 1.14 (0.05) & 2.99 (0.53) & 2.63 (0.73) & 2.49 (0.89) \\
                                                       & $10^{-4}$ & 1.74 (0.02) & 1.73 (0.02) & 1.73 (0.01) & 1.14 (0.13) & 1.12 (0.11) & 1.14 (0.34) & 3.08 (0.57) & 2.79 (0.53) & 3.03 (0.57) \\ 
                                                      & $10^{-5}$ & 1.77 (0.09) & 1.73 (0.06) & 1.74 (0.09) & 1.14 (0.16) & 1.17 (0.34) & 1.13 (0.15) & 2.88 (0.57) & 2.83 (0.44) & 3.01 (0.54) \\ 
                                                      & $10^{-6}$ & 1.78 (0.09) & 1.73 (0.04) & 1.78 (0.01) & 1.14 (0.26) & 1.17 (0.36) & 1.14 (0.24) & 2.88 (0.63) & 3.01 (0.22) & 2.90 (0.42) \\ 
                                                       & $10^{-7}$ & 1.76 (0.07) & 1.74 (0.02) & 1.78 (0.05) & 1.15 (0.29) & 1.17 (0.26) & 1.13 (0.35) & 2.97 (0.41) & 2.83 (0.56) & 3.04 (0.43) \\ 
                                                       & $10^{-8}$ & 1.73 (0.25) & 1.73 (0.11) & 1.79 (0.05) & 1.17 (0.27) & 1.17 (0.29) & 1.14 (0.07) & 2.87 (0.27) & 3.04 (0.64) & 3.01 (0.50) \\
\hline
\end{tabular}
\caption{The largest Lyapunov exponent and its standard deviation in parenthesis,  for the input sequence and for each one of the reconstructed ones as a function of the regularization parameter, $\alpha$, for the three systems.}
\label{table:LLEsVsAlpha}
\end{table}

\subsection{Latent space dimension as a function of  the size of the input sequence, $W$}

\begin{figure}[h]
\includegraphics[width = 16cm]{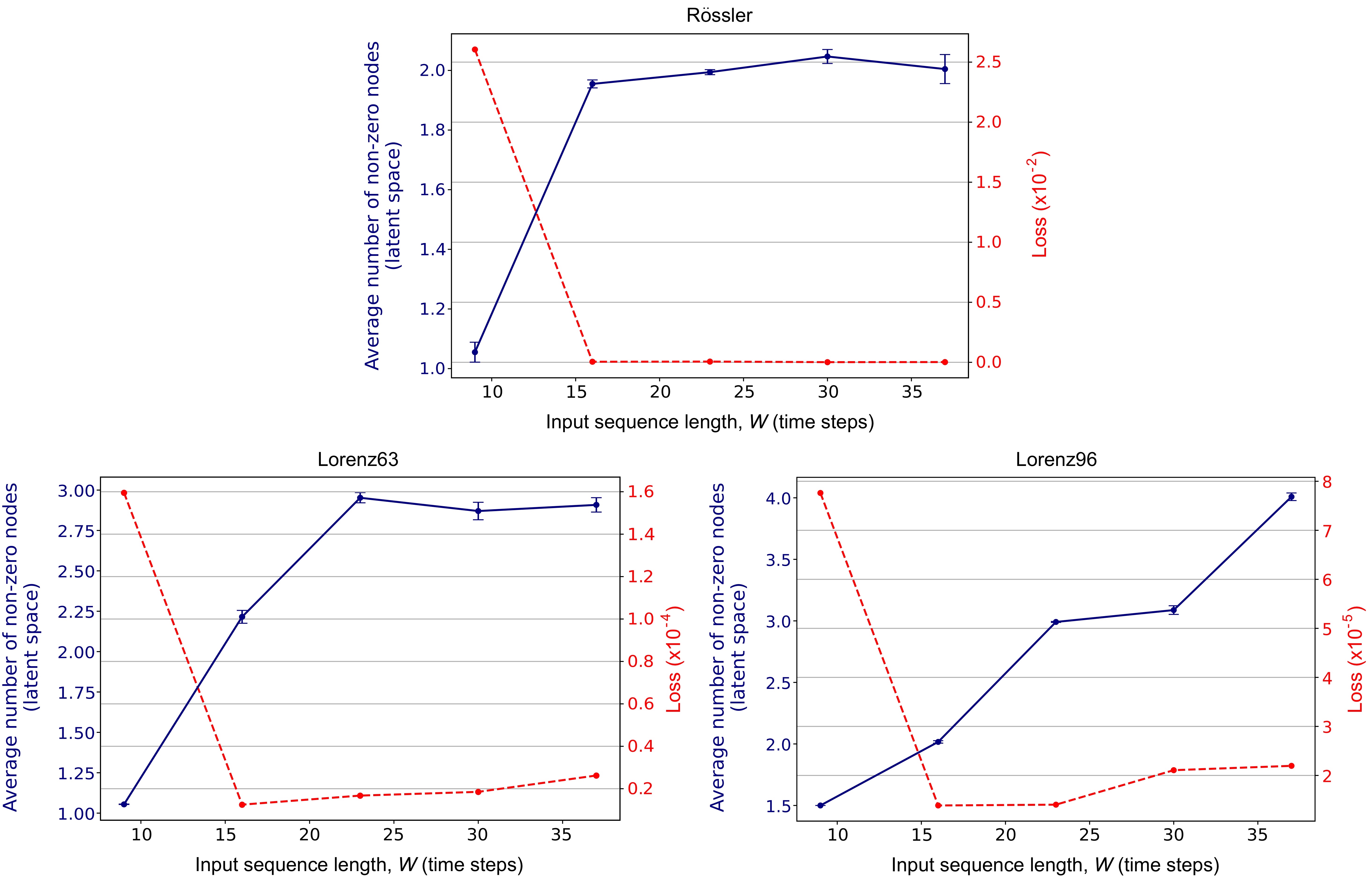}
\caption{The average number of non-zero nodes at the latent space (navy blue solid line), including the standard deviation of it (error bars) and the corresponding value of the loss function (red dashed line) of the test data as a function of the size of the input sequence, $W$, for the three systems.}
\label{latentSpaceVsWindowSize}
\end{figure}

It is important to know how the size of the information passed to the autoencoder (size of input sequence) affects the performance of the model.  In Fig.  \ref{latentSpaceVsWindowSize}, we present the average number of non-zero nodes in the latent space that were used to reconstruct the input sequence as a function of the size of the input sequence, $W$.  We fix the regularization parameter $\alpha$ to $10^{-5}$ and range the size of the input sequence, $W$ from 9 to 37 time steps using steps of 7 time steps.

In the case of the R\"ossler system, the autoencoder needs two non-zeros nodes in the latent space to reconstruct a sequence of more than 16 time steps.  Three non-zero nodes are needed in the Lorenz63 model to reconstruct a sequence of 23 time steps. We observe again, that the Lorenz96 system is more complex than the other two, and the adequate number of nodes to represent it depends strongly on the desired accuracy.  In Table \ref{table:LLEsVsWindow}, we present the LLE and its standard deviation in parenthesis for the input sequence, and the reconstructed sequence of each one of the values of the size of the input sequence, $W$.  The error in the average value of LLE found to be less than 5\% in all cases. 

\begin{table}[h!]
\centering
\begin{tabular}{c|c|c|c|c|c|c|c|c|c|c|}
& & \multicolumn{3}{c|}{R\"ossler} & \multicolumn{3}{c|}{Lorenz63} & \multicolumn{3}{c|}{Lorenz96} \\
\hline
Sequence & $W$ & X & Y & Z & X & Y & Z & 19 & 20 & 21\\
\hline
Input & - & 1.73 (0.08) & 1.72 (0.05) & 1.74 (0.09) & 1.17 (0.30) & 1.17 (0.35) & 1.14 (0.18) & 3.02 (0.88) & 3.03 (0.35) & 3.06 (0.57) \\
\hline
\multirow{6}{*}{Reconstructed} & 9 & 1.77 (0.02) & 1.75 (0.04) & 1.70 (0.19) & 1.15 (0.37) & 1.17 (0.34) & 1.15 (0.16) & 2.92 (0.40) & 2.92 (0.64) & 3.95 (0.51) \\
                                                     & 16 & 1.73 (0.21) & 1.72 (0.34) & 1.80 (0.12) & 1.13 (0.18) & 1.13 (0.34) & 1.13 (0.40) & 2.95 (0.64) & 3.05 (0.57) & 3.08 (0.21) \\
                                                     & 23 & 1.73 (0.21) & 1.76 (0.08) & 1.67 (0.44) & 1.14 (0.40) & 1.15 (0.32) & 1.15 (0.51) & 2.98 (0.39) & 2.92 (0.59) & 3.08 (0.38) \\
                                                     & 30 & 1.76 (0.14) & 1.73 (0.05) & 1.73 (0.85) & 1.14 (0.29) & 1.18 (0.13) & 1.18 (0.44) & 	2.94 (0.52) & 2.97 (0.76) & 2.92 (0.41) \\
                                                     & 37 & 1.74 (0.16) & 1.74 (0.13) & 1.82 (0.79) & 1.14 (0.18) & 1.16 (0.44) & 1.17 (0.41) & 	2.96 (0.35) & 3.03 (0.36) & 3.02 (0.46) \\
\hline
\end{tabular}
\caption{The largest Lyapunov exponent and its standard deviation in parenthesis,  for the input sequence and for each one of the reconstructed ones as a function of the size of the input sequence, $W$,  for the three systems.}
\label{table:LLEsVsWindow}
\end{table}

\section{Conclusions}

In this work, we attempted to connect chaotic dynamical systems with neural network-based autoencoders and find the minimal set of NN variables that may reproduce these dynamical systems. The construction of chaotic autoencoders is performed through deep learning networks, and in the process of their use, an information bottleneck is formed.  The latter is determined exclusively by the dimension of the latent space.  We aimed to find the minimal dimension of the latent space that faithfully reproduces the chaotic systems we studied.  This dimension depends on the required accuracy at the output.  Furthermore, it also depends on the window size we choose in order to perform the attractor reconstruction.  The analysis of the latent space dimensionality as a function of the regularization parameter and the input sequence size gives compatible numbers and this feature adds to the consistency of the picture.  In addition, the resulting maximal Lyapunov exponent of the chaotic autoencoder is respectively very close to each of the chaotic systems we analyzed.  This is a significant outcome since it shows that the chaotic autoencoder can represent faithfully complex, irregular dynamics.

An important issue that opens up from the present work is the connection of the dimensionallity of the latent space to that of the embedding dimension of the time series.  Intuitively one might think that the latent space is in some sense the minimal space required to represent a chaotic dynamical system. As a result, the latent space dimensions should be smaller or at least equal to the embedding dimension since the latter gives the framework space for analyzing the chaotic series.  We observe that this feature is true in the examples we addressed since we find latent space dimension smaller or equal to the number of differential equations that generate respective data.  An additional appealing outcome of the present analysis is that the largest Lyapunov exponent that is connected with the autoencoder system appears to be very close to the one determined directly from the dynamical system.  Clearly, these two exponents are evaluated differently since the chaotic autoencoder cannot by construction generate its dynamics and depends on the input time series.  The Lyapunov exponent, in this case, is obtained by observing the proximity of two nearby chaotic series with a very small difference in their initial conditions.  However, the autoencoder can generate similar divergence in the difference of the two series as that of the original system shows that it provides a faithful representation of the original system.  In other words, we find that the chaotic autoencoder provides a faithful representation of time series segments and the more intricate internal dynamics of the chaotic system.

The present work opens up an information window in dynamic systems.  It demonstrates that AI methods, when applied properly, may give information on complex dynamical systems and their various reduced representations.  This information is both quantitative and provides additional representations of dynamical systems.  In some sense, it also shows that neural networks are ubiquitous and may capture accurately a high degree of information contained in complex dynamical systems.  This result treated a bit more philosophically may be seen as a rather optimistic one for us since it demonstrates that human neural networks are indeed capable of capturing the complexity of the outside world.

\begin{acknowledgments}
We acknowledge the co-financing of this research by the European Union and Greek national funds through the Operational Program Crete 2020-2024, under the call "Partnerships of Companies with Institutions for Research and Transfer of Knowledge in the Thematic Priorities of RIS3Crete", with project title “Analyzing urban dynamics through monitoring the city magnetic environment” (project KPHP1 - 0029067) and also by the Ministry of Science and Higher Education of the Russian Federation in the framework of Increase Competitiveness Program of NUST "MISiS" (No. K2-2019-010), implemented by a governmental decree dated 16th of March 2013, N 211. 
\end{acknowledgments}

\section*{Data Availability Statement}
The data that support the findings of this study are available from the corresponding author upon reasonable request.


\end{document}